\documentclass{llncs}

\usepackage{times}
\usepackage{latexsym}
\usepackage{alltt}

\def\smallromani{\renewcommand{\theenumi}{\roman{enumi}}
\renewcommand{\labelenumi}{(\theenumi)}}

\newcommand{\ES}{\mbox{$\emptyset$}}

\newcommand{\A}{\mbox{$\ \wedge\ $}}

\newcommand{\sse}{\mbox{$\:\subseteq\:$}}

\newcommand{\po}{\mbox{$\ \sqsubseteq\ $}}

\newcommand{\fa}{\mbox{$\forall$}}
\newcommand{\te}{\mbox{$\exists$}}

\newcommand{\LL}{\mbox{$\ldots$}}

\newcommand{\C}[1]{\mbox{$\{{#1}\}$}}           

\newcommand{\NI}{\noindent}
\newcommand{\HB}{\hfill{$\Box$}}
\newcommand{\VV}{\vspace{5 mm}}
\newcommand{\III}{\vspace{3 mm}}
\newcommand{\II}{\vspace{2 mm}}

\newcommand{\szkew}[1]{\relax \setbox0=\hbox{\kern -24pt $\displaystyle#1$\kern 0pt }%
\box0}
{\catcode`\@=11 \global\let\ifjusthvtest@=\iffalse}

\newcounter{oldmycaption}

\newcommand{\p}[2]{\langle #1 \ ; \ #2 \rangle}

\newcommand{\Proof}{\NI
                    {\bf Proof.}\ }
\newcommand{\noprint}[1]{}
\newcommand{\noprintpb}[1]{}

\title{The Rough Guide to Constraint Propagation}

\author{Krzysztof R. Apt\inst{1,2}}
\institute{CWI\\
P.O. Box 94079, 1090 GB Amsterdam, the Netherlands\\
\email{K.R.Apt@cwi.nl}\\
\and 
University of Amsterdam, the Netherlands}

\begin{document}

\maketitle

\begin{abstract}
We provide here a simple, yet very general framework that allows us to
explain several constraint propagation algorithms in a systematic way.
In particular, using the notions commutativity and semi-commutativity,
we show how the well-known {\tt AC-3}, {\tt PC-2}, {\tt DAC} and {\tt
DPC} algorithms are instances of a single generic algorithm.  The work
reported here extends and simplifies that of Apt \cite{Apt99b}.
\end{abstract}

\section{Introduction}

Constraint programming in a nutshell consists of formulating and
solving so-called constraint satisfaction problems.  One of the most
important techniques developed in this area is constraint propagation
that aims at reducing the search space while maintaining equivalence.

We call the corresponding algorithms constraint propagation algorithms
but several other names have also been used in the literature:
consistency, local consistency, consistency enforcing, Waltz,
filtering or narrowing algorithms.  These algorithms usually aim at
reaching some form of ``local consistency'', a notion that in a
loose sense approximates the notion of ``global consistency''.

Over the last twenty few years several constraint propagation
algorithms were proposed and many of them are built into the existing
constraint programming systems.  In Apt \cite{Apt99b} we introduced a
simple framework that allows us to explain many of these algorithms in
a uniform way.  In this framework the notion of chaotic iterations, so
fair iterations of functions, on Cartesian products of specific
partial orderings played a crucial role.  In Monfroy and R{\'{e}}ty
\cite{MR99} this framework was modified to study distributed chaotic
iterations. This resulted in a general framework for distributed
constraint propagation algorithms.

We stated in Apt \cite{Apt99b} that
``the attempts of finding general principles behind the constraint
propagation algorithms repeatedly reoccur in the literature on
constraint satisfaction problems spanning the last twenty years'' and
devoted three pages to survey this work.
Two references that are perhaps closest to our work are
Benhamou \cite{Ben96} and Telerman and Ushakov \cite{TU96}. 

These developments led to an identification of a number of
mathematical properties that are of relevance for the considered
functions, namely monotonicity, inflationarity and idempotence (see,
e.g., Saraswat, Rinard and Panangaden \cite{saraswat-semantic} and
Benhamou and Older \cite{BO97}). Here we show that also the notions of
commutativity and so-called semi-commutativity are important.

As in Apt \cite{Apt99b}, to explain the constraint propagation
algorithms, we proceed here in two steps.  
First, we introduce a  generic iteration algorithm on partial
orderings and prove its correctness in an abstract setting. Then we
instantiate this algorithm with specific partial orderings and
functions.  The partial orderings will be related to the considered
variable domains and the assumed constraints, while the functions will
be the ones that characterize considered notions of local consistency
in terms of fixpoints.

This presentation allows us to clarify which properties of the
considered functions are responsible for specific properties of the
corresponding algorithms.  The resulting analysis is simpler than that
of Apt \cite{Apt99b} because we concentrate here on constraint
propagation algorithms that always terminate. This allows us to
dispense with the notion of fairness.  On the other hand, we can now
prove stronger results by taking into account the commutativity and
semi-commutativity information.

This article is organized as follows. First, in Section
\ref{sec:gen-ite}, drawing on the approach of Monfroy and R{\'{e}}ty
\cite{MR99}, we introduce a generic algorithm for the case when the
partial ordering is not further analyzed. Next, in Section
\ref{sec:compound}, we refine it for the case when the partial
ordering is a Cartesian product of component partial orderings and in
Section \ref{sec:from-to} explain how the introduced notions should be
related to the constraint satisfaction problems.

In the next four sections we instantiate the algorithm of
Section \ref{sec:gen-ite} or some of its refinements
to obtain specific constraint propagation
algorithms.  In particular, in Section 
\ref{sec:hyper-arc-algo} we derive
algorithms for arc consistency and hyper-arc
consistency.  These algorithms can be improved by taking into account
information on commutativity. This is done in Section \ref{sec:ac3}
and yields the well-known {\tt AC-3} algorithm.
Next, in Section \ref{sec:path-algo} we derive an algorithm for path
consistency and in Section \ref{sec:pc2} we improve it, again by
using information on commutativity. This yields 
the {\tt PC-2} algorithm.

In Section \ref{sec:simple-ite} we clarify under what assumptions the
generic algorithm of Section \ref{sec:gen-ite} can be simplified to a
simple {\bf for} loop statement.  Then we instantiate this simplified
algorithm to derive in Section \ref{sec:directional-arc-algo} the {\tt
  DAC} algorithm for directional arc consistency and in Section
\ref{sec:directional-path-algo} the {\tt DPC} algorithm for
directional path consistency.  Finally, in Section
\ref{sec:conclusions} we briefly discuss possible future work.

So we deal here only with the classical algorithms that establish
(directional) arc consistency and (directional) path consistency and
that are more than twenty, respectively ten, years old.  However,
several more ``modern'' constraint propagation algorithms can also be
explained in this framework. In particular, in Apt \cite[page
203]{Apt99b} we derived from a generic algorithm a simple algorithm
that achieves the notion of relational consistency of Dechter and van
Beek \cite{DvB97}.  In turn, we can use the framework of Section
\ref{sec:simple-ite} to derive the adaptive consistency algorithm of
Dechter and Pearl \cite{dechter88}.  Now, Dechter \cite{D99} showed
that this algorithm can be formulated in a very general framework of
bucket elimination that in turn can be used to explain such well-known
algorithms as directional resolution, Fourier-Motzkin elimination,
Gaussian elimination, and also various algorithms that deal with
belief networks.

Due to lack of space we do not define here formally the considered
local consistency notions and refer the interested reader instead
to the original papers or to Tsang \cite{Tsa93}.

\section{Generic Iteration Algorithms}
\label{sec:gen-ite}

Our presentation is completely general.  Consequently, we delay the
discussion of constraint satisfaction problems till Section
\ref{sec:from-to}.  In what follows we shall rely on the following
concepts.

\begin{definition}
  Consider a partial ordering $(D, \po )$ with the least element $\bot$
  and a finite set of functions $F := \C{f_1, \LL , f_k}$ on $D$.
 \begin{itemize}

\item 
By an {\em iteration of $F$\/} 
we mean an infinite sequence of values 
$d_0, d_1, \LL  $ defined inductively by
\[
d_0 := \bot,
\]
\[
d_{j} := f_{i_{j}}(d_{j-1}),
\]
where each $i_j$ is an element of $[1..k]$.

\item We say that an increasing sequence
$d_0 \: \po \: d_1 \: \po \: d_2 \: \LL$ of elements from $D$
{\em eventually stabilizes at d\/} if for some $j \geq 0$ we have
$d_i = d$ for $i \geq j$.
\HB
\end{itemize}
\end{definition}

In what follows we shall consider iterations of functions that
satisfy some specific properties.

\begin{definition}
Consider a partial ordering $(D, \po)$ and a function $f$ on $D$.

\begin{itemize}
\item $f$ is called {\em inflationary\/} \index{function!inflationary}
if $x \po f(x)$ for all $x$.

\item $f$ is called {\em monotonic\/} \index{function!monotonic}
if $x \po y$ implies 
$f(x) \po f(y)$ for all $x, y$.
\HB
\end{itemize}
\end{definition}

The following simple observation clarifies the role of monotonicity.
The subsequent result will clarify the role of inflationarity.

\begin{lemma}[Stabilization] \label{lem:stabilization}
Consider a partial ordering  $(D, \po )$ with the 
least element $\bot$
and a finite set of  monotonic functions  $F$ on $D$.

Suppose that an
iteration of $F$ eventually stabilizes at a common fixpoint $d$
of the functions from $F$.  Then $d$ is the least common fixed point
of the functions from $F$.
\end{lemma}
\Proof
Consider a common fixpoint $e$ of the functions from $F$. We prove
that $d \po e$.  Let $d_0, d_1, \LL$ be the iteration in question.
For some  $j \geq 0$ we have $d_i = d$ for $i \geq j$.

It suffices to prove by induction on $i$ that  $d_i \po e$. 
The claim obviously holds for $i = 0$ since $d_0 = \bot$.
Suppose it holds for some $i \geq 0$.
We have $d_{i+1} = f_j(d_i)$ for some $j \in [1..k]$.

By the monotonicity of $f_j$ and the induction hypothesis we get 
$f_j(d_i) \po f_j(e)$, so $d_{i+1} \po e$ since $e$ is a fixpoint of $f_j$.
\HB
\VV

We fix now a partial ordering $(D, \po )$ with the least element
$\bot$ and a set of functions $F := \C{f_1, \LL , f_k}$ on $D$.  We
are interested in computing the least common fixpoint of the functions
from $F$.  To this end we study the following algorithm
that is inspired by a similar algorithm of
Monfroy and R{\'{e}}ty \cite{MR99}.
\newpage

\NI 
{\sc   Generic Iteration Algorithm ({\tt GI})}
\begin{tabbing}
\= $d := \bot$; \\
\> $G := F$; \\ 
\> {\bf while} $G \neq \ES$ {\bf do} \\
\> \qquad choose $g \in G$; \\
\> \qquad $G := G - \C{g}$; \\
\> \qquad $G := G \cup update(G,g,d)$; \\
\> \qquad $d := g(d)$ \\
\> {\bf od} 
\end{tabbing}
where for all $G,g,d$ the set of functions 
$update(G,g,d)$ from $F$ is such that

\begin{description}
\item[A.] $\C{f \in F - G  \mid f(d) = d \A f(g(d)) \neq g(d)} \sse update(G,g,d)$,

\item[B.] $g(d) = d$ implies that $update(G,g,d) = \ES$.
  \end{description}

Intuitively, assumption {\bf A} states that $update(G,g,d)$ at least contains
all the functions from $F - G$ for which $d$ is a fixpoint but $g(d)$ is not.
The idea is that such functions are repeatedly added to the set $G$.
In turn, assumption {\bf B} states that no functions are added to $G$
in case the value of $d$ did not change.

An obvious example of an $update$ function that satisfies assumptions {\bf A}
and {\bf B} is 
\[
update(G,g,d) := \C{f \in F - G  \mid f(d) = d \A f(g(d)) \neq g(d)}.
\]
However, this choice of the $update$ function is computationally
expensive because for each function $f$ in $F - G$ we would have to
compute the values $f(g(d))$ and $f(d)$. In practice, we are interested
in some approximations of the above {\em update\/} function. We shall deal
with this matter in the next section.

We now prove correctness of this algorithm in the following sense.

\begin{theorem}[{\tt GI}] \label{thm:GI}
  \mbox{} \\[-6mm]
\begin{enumerate} \smallromani
\item Every terminating execution of the {\tt GI} algorithm computes
in $d$ a common fixpoint of the functions from $F$.

\item  Suppose that all functions in $F$ are monotonic.
Then every terminating execution of the {\tt GI} algorithm computes
in $d$ the least common fixpoint of the functions from $F$.

\item  Suppose that all functions in $F$ are inflationary and
that $(D, \po)$ is finite. Then every
  execution of the {\tt GI} algorithm terminates.
\end{enumerate}
\end{theorem}

\Proof
\II

\NI
$(i)$
Consider the predicate $I$ defined by:
\[
I :=\fa f \in F - G \  f(d) = d.
\]
Note that $I$ is established by the assignment $G := F$.  Moreover, it
is easy to check that $I$ is preserved by each {\bf while} loop
iteration.  Thus $I$ is an invariant of the {\bf while} loop of the
algorithm.  Hence upon its termination
\[
(G = \ES) \A I
\]
holds, that is  
\[
\fa f \in F \: f(d) = d.
\]

\II

\NI
$(ii)$
This is a direct consequence of $(i)$  and the
Stabilization Lemma \ref{lem:stabilization}.
\II

\NI
$(iii)$
Consider the lexicographic ordering of the partial orderings
$(D, \sqsupseteq)$ and $({\cal N}, \leq)$, 
defined on the elements of $D \times {\cal N}$ by
\[ 
(d_1, n_1) \leq_{lex} (d_2, n_2)\ {\rm iff} \ d_1 \sqsupset d_2
        \ {\rm or}\ ( d_1 = d_2 \ {\rm and}\ n_1 \leq n_2). 
\]
We use here the inverse ordering $\sqsupset$
defined by: $d_1 \sqsupset d_2$ iff $d_2 \sqsubseteq d_1$ and $d_2 \neq d_1$.

Given a finite set $G$ we denote by $card \: G$ the number of its
elements.  By assump\-tion all functions in $F$ are inflationary so, by
virtue of assumption {\bf B}, with each {\bf while} loop iteration of
the modified algorithm the pair
\[
(d, card \: G)
\]
strictly decreases in this ordering $\leq_{lex}$.
But by assumption 
$(D, \po )$ is finite, so
$(D, \sqsupseteq)$ is well-founded and consequently so is
$(D \times {\cal N}, \leq_{lex})$. This implies termination.
\HB
\VV

In particular, we obtain the following conclusion.

\begin{corollary}[{\tt GI}] \label{cor:GI}
  Suppose that $(D, \po )$ is a finite partial ordering with the least
  element $\bot$. Let $F$ be a finite set of monotonic and
  inflationary functions on $D$. Then every execution of the {\tt GI}
  algorithm terminates and computes in $d$ the least common fixpoint
  of the functions from $F$.  
\HB
\end{corollary}

In practice, we are not only interested that the {\em update\/} function
is easy to compute but also that it generates small sets of functions.
Therefore we show how the function $update$ can be made smaller when
some additional information about the functions in $F$ is available.
This will yield specialized versions of the {\tt GI} algorithm.  First
we need the following simple concepts.

\begin{definition} Consider two functions $f, g$ on a set $D$.
  \begin{itemize}
  \item 
We say that  $f$ and $g$ {\em commute\/} if 
$f (g (x)) = g (f (x))$ for all $x$.

\item We call $f$ {\em idempotent} if
$f(f(x)) = f(x)$ for all $x$.
\HB  
\end{itemize}
\end{definition}

The following result holds.

\begin{theorem}[Update] \label{thm:update} 
\mbox{} \\[-6mm]

\begin{enumerate} \smallromani

\item If $update(G,g,d)$ satisfies assumptions {\bf A} and {\bf B},
then so does the function
\[
update(G,g,d) - \C{g \mid g \mbox{ is idempotent}}.
\]

\item Suppose that for each $g \in F$ the set of functions $Comm(g)$
from $F$ is such that
\begin{itemize}

\item $g \not\in Comm(g)$,

\item each element of $Comm(g)$ commutes with $g$.

\end{itemize}
If $update(G,g,d)$ satisfies assumptions {\bf A} and {\bf B},
then so does the function
\[
update(G,g,d) - Comm(g).
\]
\end{enumerate}
\end{theorem}

\Proof
It suffices to establish in each case assumption {\bf A}.
\II

\NI
$(i)$
Suppose that $g$ is idempotent.
Then any function $f$ such that
$f(g(d)) \neq g(d)$ differs from $g$.
\II

\NI $(ii)$ Consider a function $f$ from $F- G$ such that $f(d) = d$
and $f(g(d)) \neq g(d)$. Suppose that $f \in Comm(g)$. Then $f(g(d)) =
g(f(d)) = g(d)$ which is a contradiction. So $f \not\in Comm(g)$.
Consequently, $f \in update(G,g,d) - Comm(g)$ by virtue of assumption
{\bf A} for $update(G,g,d)$.
\HB
\VV

We conclude that given an instance of the {\tt GI} algorithm that
employs a specific $update$ function, we can obtain other instances of
it by using $update$ functions modified as above.  Note that both
modifications are independent of each other and therefore can be
applied together.
In particular, when each function is idempotent
and the function $Comm$ satisfied the assumptions of $(ii)$, then
if $update(G,g,d)$ satisfies assumptions {\bf A} and {\bf B},
then so does the function $update(G,g,d) - (Comm(g) \cup \C{g})$.

\section{Compound Domains}
\label{sec:compound}

In the applications we study the iterations are carried out on a
partial ordering that is a Cartesian product of the partial orderings.
So assume now that the partial ordering $(D, \po)$ is
the Cartesian product of some partial orderings $(D_i, \po_i)$, for $i
\in [1..n]$, each with the least element $\bot_i$.  So $D = D_1 \times
\cdots \times D_n$.

 Further, we assume that each
function from $F$ depends from and affects only certain components of
$D$.  To be more precise we introduce a simple notation and
terminology.

\begin{definition} 
  Consider a sequence of partial orderings $(D_1, \po_1), \LL , (D_n,
  \po_n)$.
  \begin{itemize}

  \item By a {\em scheme\/} (on $n$) we mean a growing sequence of
different elements from $[1..n]$.  

  \item Given a scheme $s := i_1,\LL, i_l$ on $n$ we denote by 
  $(D_s, \po_s)$ the Cartesian product of the
  partial orderings $(D_{i_j}, \po_{i_j})$, for $j \in [1..l]$.

  \item Given a function $f$ on $D_s$ we say that $f$ is {\em with
      scheme $s$} and say that $f$ {\em depends on $i$\/} if $i$ is an
    element of $s$.

  \item Given an $n$-tuple $d := d_1, \LL, d_n$ from $D$ and a scheme
    $s := i_1, \LL, i_l$ on $n$ we denote by $d[s]$ the tuple
    $d_{i_1}, \LL , d_{i_l}$.  In particular, for $j \in [1..n]$ \ 
    $d[j]$ is the $j$-th element of $d$.  \HB
\end{itemize}
\end{definition} 

Consider now a function $f$ with scheme $s$.  We
extend it to a function $f^+$ from $D$ to $D$ as follows.
Take  $d \in D$. We set
\[
f^+(d) := e
\]
where  $e[s] = f(d[s])$ and $e[n-s] = d[n-s]$,
and where $n-s$ is the scheme obtained by removing from $1, \LL, n$
the elements of $s$.
We call  $f^+$ the {\em canonic extension\/} of $f$ to the domain $D$.

So $f^+(d_1, \LL, d_n) = (e_1, \LL, e_n)$ implies $d_i = e_i$
for any $i$ not in the scheme $s$ of $f$.
Informally, we can summarize it by saying that $f^+$ does not change the
components on which it does not depend. This is what we meant above by
stating that each considered function affects only certain components
of $D$.

We now say that two functions, $f$ with scheme  $s$
and $g$ with scheme  $t$ {\em commute\/} if the functions
$f^+$ and $g^+$ commute.

Instead of defining iterations for the case of the functions with
schemes, we rather reduce the situation to the one studied in the
previous section and consider, equivalently, the iterations of the
canonic extensions of these functions to the common domain $D$.
However, because of this specific form of the considered functions, we
can use now a simple definition of the $update$ function.  More
precisely, we have the following observation.

\begin{note}[Update] \label{note:update}
Suppose that each function in $F$ is of the form $f^+$. Then the 
following function $update$ satisfies assumptions {\bf A} and {\bf B}:
\begin{tabbing}
\qquad \= $update(G,g^+,d):=$ \\
       \> $\C{f^+ \in F - G \mid f \mbox{ depends on
 some } i \mbox{ in $s$ such that } d[i] \neq  g^+(d)[i]}$,
\end{tabbing}
where $g$ is with scheme $s$.
\end{note}
\Proof
To deal with assumption {\bf A} take a function
$f^+ \in F-G$ such that $f^+(d) = d$. Then 
$f(e) = e$ for any $e$ that coincides with $d$ on all components 
that are in the scheme of $f$.

Suppose now additionally that $f^+(g^+(d)) \neq g^+(d)$. By the above
$g^+(d)$ differs from $d$ on some component $i$ in the scheme of $f$.
In other words, $f$ depends on some $i$ such that $d[i] \neq
g^+(d)[i]$.  This $i$ is then in the scheme of $g$.

The proof for assumption {\bf B} is immediate.
\HB
\VV

This, together with the {\tt GI} algorithm, yields the following
algorithm in which we introduced a variable $d'$ to hold the value of
$g^+(d)$, and used $F_0 := \C{f \mid f^+ \in F}$ and the functions with
schemes instead of their canonic extensions to $D$.  
\II

\NI
{\sc Generic Iteration Algorithm for Compound Domains ({\tt CD})}
\begin{tabbing}
\= $d := (\bot_1, \LL, \bot_n)$; \\
\> $d' := d$; \\ 
\> $G := F_0$; \\ 
\> {\bf while} $G \neq \ES$ {\bf do} \\
\> \qquad choose $g \in G$; suppose $g$ is with scheme $s$; \\
\> \qquad $G := G - \C{g}$; \\
\> \qquad $d'[s] := g(d[s])$; \\
\> \qquad $G := G \cup \C{f \in  F_0 - G \mid \mbox{$f$ depends on
 some } i \mbox{ in $s$ such that } d[i] \neq d'[i]}$; \\
\> \qquad $d[s] := d'[s]$ \\
\> {\bf od} 
\end{tabbing}

The following corollary to the {\tt GI} Theorem \ref{thm:GI} and the
Update Note \ref{note:update} summarizes the correctness of this
algorithm.

\begin{corollary}[{\tt CD}] \label{cor:CD}
  Suppose that $(D, \po )$ is a finite partial ordering that is a
  Cartesian product of n partial orderings, each with the least element
  $\bot_i$ with $i \in [1..n]$. Let $F$ be a finite set of functions
  on $D$, each of the form $f^+$.

Suppose that all functions in $F$ are monotonic and inflationary. Then
every execution of the {\tt CD} algorithm terminates and computes in
$d$ the least common fixpoint of the functions from $F$.  
\HB
\end{corollary}

In the subsequent presentation we shall deal with the following
two modifications of the {\tt CD} algorithm:

\begin{itemize}
\item {\em {\tt CDI} algorithm}. This is the version of the {\tt CD} algorithm
in which all the functions are idempotent and the function $update$
defined in the Update Theorem \ref{thm:update}$(i)$ is used.

\item {\em {\tt CDC} algorithm}. This is the version of the {\tt CD}
  algorithm in which all the functions are idempotent and the combined
  effect of the functions $update$ defined in the Update Theorem
  \ref{thm:update} is used for some function $Comm$.

\end{itemize}

For both algorithms the counterparts of the {\tt CD} Corollary
\ref{cor:CD} hold.

\section{From Partial Orderings to Constraint Satisfaction Problems}
\label{sec:from-to}

We have been so far completely general in our discussion.  Recall that
our aim is to derive various constraint propagation algorithms.  To be
able to apply the results of the previous section we need to relate
various abstract notions that we used there to constraint satisfaction
problems.

This is perhaps the right place to recall the definition
and to fix the notation.
Consider a finite sequence of variables $X := x_1, \LL, x_n$,
where $n \geq 0$, with respective domains ${\cal D} := D_1, \LL, D_n$
associated with them.  So each variable $x_i$ ranges over the domain
$D_i$.  By a {\em constraint} $C$ on $X$ we mean a subset of $D_1
\times \LL \times D_n$.  

By a {\em constraint satisfaction problem}, in short CSP, we mean a
finite sequence of variables $X$ with respective domains ${\cal
  D}$, together with a finite set $\cal C$ of constraints, each on a
subsequence of $X$. We write it as $\p{{\cal C}}{x_1 \in D_1,
  \LL, x_n \in D_n}$, where $X := x_1, \LL, x_n$ and ${\cal D} :=
D_1, \LL, D_n$.

Consider now an element $d := d_1, \LL, d_n$ of $D_1 \times \LL \times
D_n$ and a subsequence $Y := x_{i_1}, \LL, x_{i_\ell}$ of
$X$. Then we denote by $d[Y]$ the sequence
$d_{i_1}, \LL, d_{i_{\ell}}$.

By a {\em solution\/} to  $\p{{\cal C}}{x_1 \in D_1, \LL, x_n \in D_n}$
we mean an element $d \in D_1 \times \LL \times D_n$ such that for
each constraint $C \in {\cal C}$ on a sequence of variables $Y$
we have $d[Y] \in C$.
We call a CSP {\em consistent\/} if it has a solution.  Two CSP's
${\cal P}_1$ and ${\cal P}_2$ with the same sequence of variables are
called {\em equivalent\/} if they have the same set of solutions.
This definition extends in an obvious way to the case of
two CSP's with the same {\em sets\/} of variables.

Let us return now to 
the framework of the previous section. It involved:
\begin{enumerate}\smallromani

\item Partial orderings with the least elements;

These will correspond to partial orderings on the CSP's.
In each of them the original CSP will be the least element
and the partial ordering will be determined by the local
consistency notion we wish to achieve.

\item Monotonic and inflationary functions with schemes;

These will correspond to the functions that transform the variable
domains or the constraints. Each function will be associated with one
or more constraints.

\item Common fixpoints;

These will correspond to the CSP's that satisfy the considered
notion of local consistency. 
\end{enumerate}

In what follows we shall discuss two specific partial orderings on the
CSP's. In each of them the considered CSP's will be defined on the
same sequences of variables.

We begin by fixing for each set $D$ a collection ${\cal F}(D)$ of the 
subsets of $D$ that includes $D$ itself. So ${\cal F}$ is a function
that given a set $D$ yields a set of its subsets to which $D$ belongs.

When dealing with the hyper-arc consistency ${\cal F}(D)$ will be simply the
set ${\cal P}(D)$ of all subsets of $D$ but for specific domains only
specific subsets of $D$ will be chosen.  For example, to deal with the
the constraint propagation for the linear constraints on integer
interval domains we need to choose for ${\cal F}(D)$ the set of all
subintervals of the original interval $D$.

When dealing with the path consistency, for a constraint $C$ the
collection ${\cal F}(C)$ will be also the set ${\cal P}(C)$ of all
subsets of $C$. However, in general other choices may be needed.  For
example, to deal with the cutting planes method, we need to limit our
attention to the sets of integer solutions to finite sets of linear
inequalities with integer coefficients (see Apt \cite[pages
193-194]{Apt99b}).

Next, given two CSP's, $\phi := \p{{\cal C}}{x_1 \in D_1, \LL, x_n \in D_n}$
and $\psi := \p{{\cal C'}}{x_1 \in D'_1, \LL, x_n \in D'_n}$, we write
$\phi \sqsubseteq_d \psi$ iff 
\begin{itemize}

\item $D'_i \in {\cal F}(D_i)$ (and hence $D'_i \sse D_i$)
for $i \in [1..n]$,

\item the constraints in ${\cal C'}$ are the restrictions of the
  constraints in ${\cal C}$ to the domains $D'_1, \LL, D'_n$.
\end{itemize}

So $\phi \sqsubseteq_d \psi$ if $\psi$ can be obtained from $\phi$ by a
domain reduction rule and the domains of $\psi$ belong to the
appropriate collections of sets ${\cal F}(D)$.

Next, given two CSP's, $\phi := \p{C_1, \LL, C_k}{{\cal DE}}$
and $\psi := \p{C'_1, \LL, C'_k}{{\cal DE}}$, we write
$\phi \sqsubseteq_c \psi$ iff 

\begin{itemize}

\item $C'_i \in {\cal F}(C_i)$ (and hence $C'_i \sse C_i$)
for $i \in [1..k]$.

\end{itemize}

In what follows we call $\sqsubseteq_d$ the {\em domain reduction
  ordering\/} and $\sqsubseteq_c$ the {\em constraint reduction
  ordering}.  To deal with the arc consistency, hyper-arc consistency
and directional arc consistency notions we shall use the domain
reduction ordering, and to deal with path consistency and directional
path consistency notions we shall use the constraint reduction
ordering.

We consider each ordering with some fixed initial CSP ${\cal P}$
as the least element. In other words, each domain reduction ordering
is of the form
\[
(\C{{\cal P'} \mid {\cal P} \sqsubseteq_d {\cal P'}}, \sqsubseteq_d)
\]
and each constraint reduction ordering is of the form
\[
(\C{{\cal P'} \mid {\cal P} \sqsubseteq_c {\cal P'}}, \sqsubseteq_c).
\]

Note that 
$
\p{{\cal C}}{x_1 \in D'_1, \LL, x_n \in D'_n} \sqsubseteq_d
\p{{\cal C'}}{x_1 \in D''_1, \LL, x_n \in D''_n}$
iff
$D'_i \supseteq D''_i \mbox{ for } i \in [1..n]$.

This means that for
${\cal P} = \p{{\cal C}}{x_1 \in D_1, \LL, x_n \in D_n}$
we can identify the domain reduction ordering
$(\C{{\cal P'} \mid {\cal P} \sqsubseteq_d {\cal P'}}, \sqsubseteq_d)$
with the Cartesian product of the partial orderings 
$({\cal F}(D_i), \supseteq)$, where $i \in [1..n]$.
Additionally, each CSP in this domain reduction ordering is uniquely
determined by its domains and by the initial ${\cal P}$.

Similarly, 
\[
\p{C'_1, \LL, C'_k}{{\cal DE}} \sqsubseteq_c  \p{C''_1, \LL, C''_k}{{\cal DE}} \mbox{ iff }
C'_i  \supseteq C''_i \mbox{ for } i \in [1..k].
\]
This allows us for ${\cal P} = \p{C_1, \LL, C_k}{{\cal DE}}$
to identify the constraint reduction ordering
$(\C{{\cal P'} \mid {\cal P} \sqsubseteq_c {\cal P'}}, \sqsubseteq_c)$
with the Cartesian product of the partial orderings 
$({\cal F}(C_i), \supseteq)$, where $i \in [1..k]$.
Also, each CSP in this constraint reduction ordering is uniquely
determined by its constraints and by the initial ${\cal P}$.

In what follows instead of the domain reduction ordering and the
constraint reduction ordering we shall use the corresponding Cartesian
products of the partial orderings.  So in these compound orderings the
sequences of the domains (respectively, of the constraints) are
ordered componentwise by the reversed subset ordering
$\supseteq$. Further, in each component ordering $({\cal F}(D),
\supseteq)$ the set $D$ is the least element.

Consider now a function $f$ on some Cartesian product
${\cal F}(E_1) \times \LL \times {\cal F}(E_m)$.
Note that $f$ is inflationary w.r.t. the componentwise ordering $\supseteq$ if 
for all $(X_1, \LL, X_m) \in {\cal F}(E_1) \times \LL \times {\cal F}(E_m)$
we have $Y_i \subseteq X_i$ for all $i \in [1..m]$, where 
$f(X_1, \LL, X_m) = (Y_1, \LL, Y_m)$.

Also, $f$ is monotonic w.r.t. the componentwise ordering $\supseteq$ if for all 
$(X_1, \LL, X_m),$ $(X'_1, \LL, X'_m) \in {\cal F}(E_1) \times \LL \times {\cal F}(E_m)$
such that $X_i \subseteq X'_i$ for all $i \in [1..m]$,
the following holds: if
\[
\mbox{$f(X_1, \LL, X_m) = (Y_1, \LL, Y_m)$ and
$f(X'_1, \LL, X'_m) = (Y'_1, \LL, Y'_m)$,}
\]
then
$Y_i \subseteq Y'_i$ for all $i \in [1..m]$. 

In other words, $f$ is monotonic w.r.t.
$\supseteq$ iff it is monotonic w.r.t. $\subseteq$.
This reversal of the set inclusion of course does not hold
for the inflationarity notion.

\section{A Hyper-arc Consistency Algorithm}
\label{sec:hyper-arc-algo}

We begin by considering the notion of hyper-arc consistency of Mohr
and Masini \cite{MM88} (we use here the terminology of 
Marriott and Stuckey \cite{MS98b}).
The more known notion of arc consistency of Mackworth
\cite{mackworth-consistency} is obtained by restricting one's
attention to binary constraints.

To employ the {\tt CDI} algorithm of Section \ref{sec:compound}
we now make specific choices involving the items (i), (ii) and
(iii) of the previous section.
\II

\NI
Re: (i) Partial orderings with the least elements.

As already mentioned in the previous section, for the function ${\cal
  F}$ we choose the powerset function ${\cal P}$, so for each domain
$D$ we put ${\cal F}(D) := {\cal P}(D)$.

Given a CSP ${\cal P}$ with the sequence $D_1, \LL, D_n$ of the
domains we take the domain reduction ordering with ${\cal P}$ as its
least element. As already noted we can identify this ordering with the
Cartesian product of the partial orderings $({\cal P}(D_i),
\supseteq)$, where $i \in [1..n]$.  The elements of this compound
ordering are thus sequences $(X_1, \LL, X_n)$ of respective subsets of
the domains $D_1, \LL, D_n$ ordered componentwise by the reversed
subset ordering $\supseteq$.
\II

\NI
Re: (ii) Monotonic and inflationary functions with schemes.

Given a constraint $C$ on the variables $y_1, \LL, y_k$ with
respective domains $E_1, \LL, E_k$, we abbreviate for each $j \in
[1..k]$ the set $\C{d[j] \mid d \in C}$ to $\Pi_{j}(C)$.  Thus
$\Pi_{j}(C)$ consists of all $j$-th coordinates of the elements of
$C$.  Consequently, $\Pi_{j}(C)$ is a subset of the domain $E_j$ of
the variable $y_j$.

We now introduce for each $i \in [1..k]$
the following function
$\pi_i$ on ${\cal P}(E_1) \times \cdots \times {\cal P}(E_k)$:

\[
\pi_i(X_1, \LL , X_k) := (X_1, \LL,  X_{i-1}, X'_i, X_{i+1}, \LL, X_k)
\]
where 
\[
X'_i := \Pi_i(C \cap (X_1 \times \cdots \times X_k)).
\]
That is, 
$X'_i = \C{d[i] \mid d \in X_1 \times \cdots \times X_k \mbox { and } d \in C}$.
Each function $\pi_i$ is associated with a specific constraint $C$.
Note that $X'_i \sse X_i$, so each function $\pi_i$ boils down to a
projection on the $i$-th component.  
\II

\NI
Re: (iii) Common fixpoints.

Their use is clarified by the following lemma that also lists the
relevant properties of the functions $\pi_i$.

\begin{lemma}[Hyper-arc Consistency] \label{lem:hyper-arc}
 \mbox{} \\[-6mm]
  \begin{enumerate}\smallromani
  \item A CSP $\p{{\cal C}}{x_1 \in D_1, \LL, x_n \in D_n}$
is hyper-arc consistent iff $(D_1, \LL, D_n)$ is a
common fixpoint of all functions $\pi^{+}_i$ associated 
with the constraints from ${\cal C}$.

  \item
Each projection function $\pi_i$ associated with a constraint $C$ is 
\begin{itemize}
\item inflationary w.r.t. the componentwise  ordering $\supseteq$,

\item monotonic w.r.t. the componentwise  ordering $\supseteq$,

\item idempotent.
\HB
\end{itemize}
\end{enumerate}
\end{lemma}

By taking into account only the binary constraints we obtain an
analogous characterization of arc consistency.  The functions $\pi_1$
and $\pi_2$ can then be defined more directly as follows:
\[
\pi_{1}(X,Y) := (X',Y),
\]
where
$X':= \C{a \in X \mid \te \: b \in Y \: (a,b) \in C}$,
and
\[
\pi_2(X, Y) := (X,Y'),
\]
where $Y' := \C{b \in Y \mid \te a \in X \: (a,b) \in C}$.

Fix now a CSP ${\cal P}$. By instantiating the {\tt CDI} algorithm with 
\[
F_0 := \C{f \mid f \mbox{ is a $\pi_i$ function associated with a
constraint of ${\cal P}$}}
\] 
and with each $\bot_i$ equal to $D_i$ we get the {\tt
HYPER-ARC} algorithm that enjoys following properties.
 
\begin{theorem}[{\tt HYPER-ARC} Algorithm] \label{thm:hyper-arc}
Consider a CSP
${\cal P} := \p{{\cal C}}{x_1 \in D_1, \LL, x_n \in D_n}$
where each $D_i$ is finite.

The {\tt HYPER-ARC} algorithm always terminates.
Let ${\cal P'}$ be the CSP determined by ${\cal P}$ and
the sequence of the domains $D'_1, \LL,  D'_n$ 
computed in $d$. Then
\begin{enumerate}\smallromani

\item ${\cal P'}$ is the $\sqsubseteq_d$-least CSP that is hyper-arc consistent,

\item ${\cal P'}$ is equivalent to ${\cal P}$.
\HB
\end{enumerate}
\end{theorem}

Due to the definition of the $\sqsubseteq_d$ ordering the item $(i)$ can be
rephrased as follows.  Consider all hyper-arc consistent CSP's that are of
the form $\p{{\cal C'}}{x_1 \in D'_1, \LL, x_n \in D'_n}$ where $D'_i
\sse D_i$ for $i \in [1..n]$ and the constraints in ${\cal C'}$ are
the restrictions of the constraints in ${\cal C}$ to the domains
$D'_1, \LL, D'_n$.  Then among these CSP's ${\cal P'}$ has the largest
domains.

\section{An Improvement: the {\tt AC-3} Algorithm}
\label{sec:ac3}

In this section we show how we can exploit an information about the
commutativity of the $\pi_i$ functions.  Recall
that in Section \ref{sec:compound} we modified the notion of
commutativity for the case of functions with schemes. We now need the
following lemma.

\begin{lemma}[Commutativity] \label{lem:comm}
Consider a CSP and two constraints of it, $C$ on the variables
$y_1, \LL, y_k$ and $E$ on the variables $z_1, \LL, z_{\ell}$.
\begin{enumerate}\smallromani

\item For $i,j \in [1..k]$ the functions  $\pi_i$ and $\pi_j$ 
of the constraint $C$ commute.

\item If the variables $y_i$ and $z_j$ are identical then the functions $\pi_i$
of $C$ and $\pi_j$ of $E$ commute.
\HB
\end{enumerate} 
\end{lemma}

Fix now a CSP. We derive a modification of the {\tt HYPER-ARC}
algorithm by instantiating this time the {\tt CDC} algorithm. As 
before we use the set of functions
$
F_0 := \C{f \mid f \mbox{ is a $\pi_i$ function associated with a
constraint of ${\cal P}$}}
$
and each $\bot_i$ equal to $D_i$.
Additionally we employ the following function {\em Comm\/},
where $\pi_i$ is associated with a constraint $C$:

\begin{tabbing}
\quad $Comm(\pi_i)$ \=  := \= \C{\pi_j \mid} \kill 
\quad $Comm(\pi_i)$ \>  := \> \C{\pi_j \mid 
\mbox{$i \neq j $ and $\pi_j$ is associated with the constraint $C$}} \\
\quad             \> $\cup$ \> \{$\pi_j \mid$
\mbox{$\pi_j$ is associated with a constraint $E$ and} \\
            \>        \> \mbox{ \ $\quad$ \ the $i$-th variable of $C$ and the $j$-th variable of $E$ coincide\}}.
\end{tabbing}

By virtue of the Commutativity Lemma \ref{lem:comm} 
each set $Comm(g)$ satisfies the assumptions of the Update Theorem
\ref{thm:update}$(ii)$.

By limiting oneself to the set of functions $\pi_1$ and $\pi_2$
associated with the binary constraints,
we obtain an analogous modification of the 
corresponding arc consistency algorithm.

Using now the counterpart of the {\tt CD} Corollary \ref{cor:CD} for
the {\tt CDC} algorithm we conclude that the above algorithm enjoys
the same properties as the {\tt HYPER-ARC} algorithm, that is the
counterpart of the {\tt HYPER-ARC} Algorithm Theorem
\ref{thm:hyper-arc} holds.

Let us clarify now the difference between this algorithm and the {\tt
HYPER-ARC} algorithm when both of them are limited to the binary
constraints.

Assume that the considered CSP is of the form $\p{{\cal C}}{{\cal DE}}$.
We reformulate the above algorithm as follows.
Given a binary relation $R$, we put
\[
R^{T} := \C{(b,a) \mid (a,b) \in R}.
\]

For $F_0$ we now choose the set of the $\pi_1$ functions of the 
constraints or relations from the set
\begin{tabbing}
\qquad \= $S_0$ \= :=     \= \C{C \mid \mbox{$C$ is a binary constraint from ${\cal C}$}} \\
       \>       \> $\cup$ \> \C{C^{T} \mid \mbox{$C$ is a binary constraint from ${\cal C}$}}.
\end{tabbing}

Finally, for each $\pi_1$ function of some $C \in S_0$ on $x,y$ 
we define
\begin{tabbing}
\qquad  $Comm(\pi_1)$ \=  :=  \= \C{\pi_j \mid} \kill
\qquad $Comm(\pi_1)$ \>  :=  \> \C{f \mid \mbox{$f$ is the $\pi_1$ function of $C^{T}$}} \\
\qquad               \> $\cup$ \> \C{f \mid \mbox{$f$ is the $\pi_1$ function of some $E \in S_0$ on
                         $x,z$ where $z \not \equiv y$}}.
\end{tabbing}

Assume now that
\begin{equation}
  \label{eq:atmost}
\mbox{for each pair of variables $x,y$ at most one constraint exists on $x,y$.}  
\end{equation}

Consider now the corresponding instance of the {\tt CDC} algorithm.
By incorporating into it the effect of the functions $\pi_1$ on the
corresponding domains, we obtain the following
algorithm known as the {\tt AC-3} algorithm of Mackworth
\cite{mackworth-consistency}.

We assume here that ${\cal DE} := x_1 \in D_1, \LL, x_n \in D_n$.
\II

\NI
{\sc {\tt AC-3\/} Algorithm}

\begin{tabbing}
\= $S_0$ \= := \= \C{C \mid \mbox{$C$ is a binary constraint from ${\cal C}$}} \\
       \>       \> $\cup$ \> \C{C^{T} \mid \mbox{$C$ is a binary constraint from ${\cal C}$}}; \\
\> $S := S_0$; \\ 
\> {\bf while} $S \neq \ES$ {\bf do} \\
\> \qquad choose $C \in S$; suppose $C$ is on $x_i, x_j$; \\
\> \qquad $D_i := \C{a \in D_i \mid \te \: b \in D_j \: (a,b) \in C}$; \\
\> \qquad {\bf if} $D_i$ changed {\bf then} \\
\> \qquad \qquad $S := S \cup \C{ C' \in S_0 \mid C' 
  \mbox{ is on the variables  $y, x_i$ where $y \not \equiv x_j$} }$ \\
\> \qquad {\bf fi}; \\
\> \qquad $S := S - \C{C}$ \\
\> {\bf od}
\end{tabbing}
\III

It is useful to mention that the corresponding reformulation of the
{\tt HYPER-ARC} algorithm differs in the second assignment to $S$
which is then
\[
S := S \cup \C{ C' \in S_0 \mid C' \mbox{ is on the variables  $y, z$ where $y$ is $x_i$ or $z$ is $x_i$}}.
\]

So we ``capitalized'' here on the commutativity of the corresponding
projection functions $\pi_1$ as follows. First, no constraint or
relation on $x_i, z$ for some $z$ is added to $S$.  Here we exploited
part $(ii)$ of the Commutativity Lemma \ref{lem:comm}.

Second, no constraint or relation on $x_j, x_i$ is added to $S$.  Here
we exploited part $(i)$ of the Commutativity Lemma \ref{lem:comm},
because by assumption (\ref{eq:atmost}) $C^{T}$ is the
only constraint or relation on $x_j, x_i$ and its $\pi_1$ function
coincides with the $\pi_2$ function of $C$.

In case the assumption (\ref{eq:atmost})
about the considered CSP is dropped, the resulting algorithm
is somewhat less readable. However, once
we use the following
modified definition of $Comm(\pi_1)$:
\[
Comm(\pi_1) := \C{f \mid \mbox{$f$ is the $\pi_1$ function of some $E \in S_0$ on $x,z$ where $z \not \equiv y$}}
\]
we get an instance of the {\tt CDC}
algorithm which differs from the {\tt AC-3} algorithm in
that the qualification ``where $y \not \equiv x_j$'' is removed from the
definition of the second assignment to the set $S$.

\section{A Path Consistency Algorithm}
\label{sec:path-algo}

The notion of path consistency was introduced in Montanari
\cite{montanari-networks}.  It is defined for special type of CSP's.
For simplicity we ignore here unary constraints that are usually
present when studying path consistency.

\begin{definition}
  We call a CSP {\em normalized\/} if it has only binary constraints
and for each pair $x,y$ of its variables exactly one constraint on
them exists. We denote this constraint by $C_{x,y}$.  
\HB
\end{definition}

Every CSP with only unary and binary constraints
is trivially equivalent to a normalized CSP.
Consider now a normalized CSP ${\cal P}$.
Suppose that ${\cal P} = \p{{C_1, \LL, C_k}}{{\cal DE}}$.

We proceed now as in the case of hyper-arc consistency.  First, we
choose for the function ${\cal F}$ the powerset function.  For
the partial ordering we choose the constraint reduction ordering
of Section \ref{sec:from-to}, or rather its counterpart which is the
Cartesian product of the partial orderings $({\cal P}(C_i),
\supseteq)$, where $i \in [1..k]$.

Second, we introduce appropriate monotonic and inflationary functions with schemes.
To this end, given two binary relations $R$ and $S$
we define their composition $\cdot$ by
\[
R \cdot S :=   \C{(a,b) \mid \te c \: ((a,c) \in R, (c,b) \in S)}.
\]

Note that if $R$ is a constraint on the variables $x,y$ and $S$ a
constraint on the variables $y,z$, then $R \cdot S$ is a
constraint on the variables $x,z$.

Given a subsequence $x,y,z$ of the variables of ${\cal P}$ we now
introduce three functions on ${\cal P}(C_{x,y}) \times {\cal
  P}(C_{x,z}) \times {\cal P}(C_{y,z})$:

\[
f^{z}_{x,y}(P,Q,R) := (P',Q,R),
\]
where $P' := P \cap Q \cdot R^T$,
\[
f^{y}_{x,z}(P,Q,R) := (P,Q',R),
\]
where $Q' := Q \cap P \cdot R$,
and
\[
f^{x}_{y,z}(P,Q,R) := (P,Q,R'),
\]
where $R' := R \cap P^T \cdot Q$.

Finally, we introduce common fixpoints of the above defined functions.
To this end we need the following counterpart of the Hyper-arc
Consistency Lemma \ref{lem:hyper-arc}.

\begin{lemma}[Path Consistency]\label{lem:path}
 \mbox{} \\[-6mm]
  \begin{enumerate}\smallromani
  \item 
A normalized CSP $\p{{C_1, \LL, C_k}}{{\cal DE}}$
is path consistent iff $(C_1, \LL, C_k)$ is a common fixpoint of all
functions $(f^{z}_{x,y})^{+}$, $(f^{y}_{x,z})^{+}$ and
$(f^{x}_{y,z})^{+}$ associated with
the subsequences $x,y,z$ of its variables.

\item
The functions  $f^{z}_{x,y}$, $f^{y}_{x,z}$ and $f^{x}_{y,z}$ are
\begin{itemize}
\item inflationary w.r.t. the componentwise  ordering $\supseteq$,

\item monotonic w.r.t. the componentwise  ordering $\supseteq$,

\item idempotent.
\HB

\end{itemize}
\end{enumerate}
\end{lemma}

We now instantiate the {\tt CDI} algorithm with the set of functions
\[
F_0 := \C{f \mid x,y,z 
\mbox{ is a subsequence of the variables of ${\cal P}$ and
$f \in \C{f^{z}_{x,y}, f^{y}_{x,z}, f^{x}_{y,z}}$}},
\]
$n := k$ and each $\bot_i$ equal to $C_i$.

Call the resulting algorithm the {\tt PATH} algorithm.
It enjoys the following properties.
 
\begin{theorem}[{\tt PATH} Algorithm] \label{thm:path-algo}
Consider a normalized CSP 
${\cal P} := \p{{C_1, \LL, C_k}}{{\cal DE}}$. 
Assume that each constraint $C_i$ is finite. 

The {\tt PATH} algorithm always terminates.  Let 
${\cal P'} := \p{{C'_1, \LL, C'_k}}{{\cal DE}}$, where
the sequence of the
constraints $C'_1, \LL, C'_k$ is computed in $d$. Then

\begin{enumerate}\smallromani

\item ${\cal P'}$ is the $\sqsubseteq_c$-least CSP 
that is path consistent,

\item ${\cal P'}$ is equivalent to ${\cal P}$.
\HB
\end{enumerate}
\end{theorem}

As in the case of the {\tt HYPER-ARC} Algorithm Theorem \ref{thm:hyper-arc}
the item  $(i)$
can be rephrased as follows.  Consider all path
consistent CSP's that are of the form $\p{{C'_1, \LL, C'_k}}{{\cal DE}}$
where $C'_i \sse C_i$ for $i \in [1..k]$.  Then among them
${\cal P'}$ has the largest constraints.

\section{An Improvement: the {\tt PC-2} Algorithm}
\label{sec:pc2}

As in the case of the hyper-arc consistency we can improve the {\tt
PATH} algorithm by taking into account the commutativity information.

Fix a normalized CSP ${\cal P}$. We abbreviate the statement ``$x,y$
is a subsequence of the variables of ${\cal P}$'' to $x \prec y$.
We now have the following lemma.

\begin{lemma}[Commutativity] \label{lem:comm-path}
Suppose that $x \prec y$ and let $z,u$ be some variables of
${\cal P}$ such that $\C{u,z} \cap \C{x,y} = \ES$.
Then the functions $f^{z}_{x,y}$ and $f^{u}_{x,y}$ commute.
\HB
\end{lemma}
In other words, two functions with the same pair of
variables as a subscript commute.
\II

We now instantiate the {\tt CDC} algorithm with the same set of
functions $F_0$ as in Section \ref{sec:path-algo}.
Additionally, we use the function $Comm$ defined as follows,
where $x \prec y$ and where $z\not\in \C{x,y}$:

\[
Comm(f^{z}_{x,y}) = \C{f^{u}_{x,y} \mid u \not\in \C{x,y,z}}.
\]

Thus for each function $g$ the set $Comm(g)$ contains precisely $m-3$
elements, where $m$ is the number of variables of the considered
CSP. This quantifies the maximal ``gain'' obtained by using the commutativity
information: at each ``update'' stage of the corresponding instance of
the {\tt CDC} algorithm we add up to $m-3$ less elements than in the case of
the corresponding instance of the {\tt CDI} algorithm considered in
the previous section.  

By virtue of the Commutativity Lemma \ref{lem:comm-path} each set
$Comm(g)$ satisfies the assumptions of the Update Theorem
\ref{thm:update}$(ii)$.  We conclude that the above instance of the
{\tt CDC} algorithm enjoys the same properties as the original {\tt
PATH} algorithm, that is the counterpart of the {\tt PATH} Algorithm
Theorem \ref{thm:path-algo} holds.  To make this modification of the
{\tt PATH} algorithm easier to understand we proceed as follows.

Each function of the form $f^{u}_{x,y}$ where $x \prec y$ and $u
\not\in \C{x,y}$ can be identified with the sequence $x,u,y$ of the
variables. (Note that the ``relative'' position of $u$ w.r.t. $x$ and
$y$ is not fixed, so $x,u,y$ does not have to be a subsequence of the
variables of ${\cal P}$.)
This allows us to identify the set of functions $F_0$ with the set
\[
V_0 := \C{(x,u,y) \mid x \prec y, u \not\in \C{x,y}}.
\]

Next, assuming that $x \prec y$,
we introduce the following set of triples of different variables of ${\cal P}$:

\begin{tabbing}
\qquad $V_{x,y}$ \= :=     \= \C{(x,y,u) \mid x \prec u} $\cup$ \C{(y,x,u) \mid y \prec u} \\
                 \> $\cup$ \> \C{(u,x,y) \mid u \prec y} $\cup$ \C{(u,y,x) \mid u \prec x}.
\end{tabbing}

Informally, $V_{x,y}$ is the subset of $V_0$ that consists of the triples
that begin or end with either $x,y$ or $y,x$.
This corresponds to the set of functions in one of the following forms:
$f^{y}_{x,u}, f^{x}_{y,u}, f^{x}_{u,y}$ and $f^{y}_{u,x}$.

The above instance of the {\tt CDC} algorithm then becomes
the following {\tt PC-2} algorithm of Mackworth
\cite{mackworth-consistency}. Here
initially $E_{x,y} = C_{x,y}$.
\II

\NI
{\sc {\tt PC-2\/} Algorithm}

\begin{tabbing}
\= $V_0$ := $\C{(x,u,y) \mid x \prec y, u \not\in \C{x,y}}$; \\
\> $V := V_0$; \\ 
\> {\bf while} $V \neq \ES$ {\bf do} \\
\> \qquad choose $p \in V$; suppose $p = (x,u,y)$; \\
\> \qquad apply  $f^{u}_{x,y}$ to its current domains; \\
\> \qquad {\bf if} $E_{x,y}$ changed {\bf then} \\
\> \qquad \qquad $V := V \cup V_{x,y}$; \\
\> \qquad {\bf fi}; \\
\> \qquad $V := V - \C{p}$ \\
\> {\bf od}
\end{tabbing}

Here the phrase ``apply  $f^{u}_{x,y}$ to its current domains''
can be made more precise if the ``relative'' position of $u$ w.r.t. $x$ and
$y$ is known. Suppose for instance that $u$ is ``before'' $x$ and $y$.
Then $f^{u}_{x,y}$ is defined on 
${\cal P}(C_{u,x}) \times {\cal P}(C_{u,y}) \times {\cal P}(C_{x,y})$ by
\[
f^{u}_{x,y}(E_{u,x},E_{u,y},E_{x,y}) := (E_{u,x},E_{u,y}, E_{x,y} \cap E_{u,x}^T \cdot E_{u,y}),
\]
so the above phrase ``apply  $f^{u}_{x,y}$ to its current domains''
can be replaced by the assignment
\[
E_{x,y} := E_{x,y} \cap E_{u,x}^T \cdot E_{u,y}.
\]
Analogously for the other two possibilities.

The difference between the {\tt PC-2\/} algorithm and the corresponding
representation of the {\tt PATH} algorithm lies in the way
the modification of the set $V$ is carried out. In the case
of the {\tt PATH} algorithm the second assignment to $V$ is
\[
V := V \cup V_{x,y} \cup \C{(x,u,y) \mid u \not\in \C{x,y}}.
\]

\section{Simple Iteration Algorithms}
\label{sec:simple-ite}

Let us return now to the framework of Section \ref{sec:gen-ite}.  We
analyze here when the {\bf while} loop of the {\sc Generic Iteration
Algorithm} {\tt GI} can be replaced by a {\bf for} loop.  First, we
weaken the notion of commutativity as follows.

\begin{definition}
Consider a partial ordering  $(D, \po )$ and functions
$f$ and $g$ on $D$.
We say that  $f$ {\em semi-commutes with\/} $g$ ({\em w.r.t. $\po$\/}) if 
$f (g (x)) \po g (f (x))$ for all $x$.
\HB  
\end{definition}

The following lemma provides an answer to the  question just posed.
Here and elsewhere we omit brackets when writing repeated applications
of functions to an argument.

\begin{lemma}[Simple Iteration] \label{lem:semi-com}            
  Consider a partial ordering $(D, \po )$ with the least element
  $\bot$. Let $F := f_1, \LL, f_k$ be a finite sequence of monotonic,
  inflationary and idempotent functions on $D$.  Suppose that 
  $f_i$ semi-commutes with $f_j$ for $i > j$, that is,
\begin{equation}
f_i(f_j(x)) \po f_j(f_i(x)) \mbox{ for all $x$.}
\label{equ:semi-com}
\end{equation}
Then $f_1 f_2 \LL f_k(\bot)$ is the least common fixpoint
of the functions from $F$.  
\HB
\end{lemma}
\Proof
We prove first that for $i \in [1..k]$ we have
\[
f_i f_1 f_2 \LL f_k(\bot) \po f_1 f_2 \LL f_k(\bot).
\]
Indeed, by the assumption (\ref{equ:semi-com}) we have the 
following string of inclusions,
where the last one is due to the idempotence of the considered functions:
\[
f_i f_1 f_2 \LL f_k(\bot) \po f_1 f_i f_2 \LL f_k(\bot) \po \LL \po
f_1 f_2 \LL f_i f_i \LL f_k(\bot) \po f_1 f_2 \LL f_k(\bot).
\]

Additionally, by the inflationarity of the considered functions,
we also have for $i \in [1..k]$
\[
f_1 f_2 \LL f_k(\bot) \po f_i f_1 f_2 \LL f_k(\bot).
\]

So $f_1 f_2 \LL f_k(\bot)$ is a common fixpoint of the functions from
$F$.  This means that the iteration of $F$ that starts with $\bot$,
$f_k(\bot)$, $f_{k-1} f_k(\bot), \LL, f_1 f_2 \LL f_k(\bot)$
eventually stabilizes at $f_1 f_2 \LL f_k(\bot)$. By the Stabilization
Lemma \ref{lem:stabilization} we get the desired conclusion.  

\HB 
\VV

The above lemma provides us with a simple way of computing the least
common fixpoint of a set of finite functions that satisfy the
assumptions of this lemma, in particular condition
(\ref{equ:semi-com}). Namely, it suffices to order these functions in
an appropriate way and then to apply each of them just once, starting
with the argument $\bot$.

To this end we maintain the considered functions not in a set but in a
list.  Given a non-empty list $L$ we denote its head by ${\bf
  head}(L)$ and its tail by ${\bf tail}(L)$.  Next, given a sequence
of elements $a_1, \LL, a_n$ with $n \geq 0$, we denote by $[a_1, \LL,
a_n]$ the list formed by them. If $n=0$, then this list is empty and is
denoted by $[ \: ]$ and if $n > 0$, then ${\bf head}([a_1, \LL, a_n])
= a_1$ and ${\bf tail}([a_1, \LL, a_n]) = [a_2, \LL a_n]$. 

The following algorithm is a counterpart of the {\tt GI} algorithm.
We assume in it that condition (\ref{equ:semi-com}) holds for
the functions $f_1, \LL, f_k$.
\II

\NI
{\sc Simple Iteration Algorithm ({\tt SI})}
\begin{tabbing}
\= $d := \bot$; \\
\> $L := [f_k, f_{k-1}, \LL, f_1]$; \\
\> {\bf for} $i := 1$ {\bf to} $k$ {\bf do} \\
\> \qquad $g := {\bf head}(L)$; \\
\> \qquad $L := {\bf tail}(L)$; \\
\> \qquad $d := g(d)$ \\
\> {\bf od} 
\end{tabbing}

The following immediate consequence of the Simple Iteration Lemma
\ref{lem:semi-com} is a counterpart of the {\tt GI} Corollary
\ref{cor:GI}.

\begin{corollary}[{\tt SI}] \label{cor:SI}
  Suppose that $(D, \po )$ is a partial ordering with the least
  element $\bot$. Let $F := f_1, \LL, f_k$ be a finite sequence of
  monotonic, inflationary and idempotent functions on $D$ such that
  (\ref{equ:semi-com}) holds. Then the {\tt SI} algorithm terminates
  and computes in $d$ the least common fixpoint of the functions from
  $F$.  \HB
\end{corollary}

Note that in contrast to the {\tt GI} Corollary
\ref{cor:GI} we do not require here that the partial ordering is finite.
Because at each iteration of the {\bf for} loop exactly one element is
removed from the list $L$,  at the end of this loop the list $L$ is
empty. Consequently, this algorithm is a reformulation
of the one in which the line  
\[
\mbox{{\bf for} $i := 1$ {\bf to} $k$ {\bf do}}
\]
is replaced by 
\[
\mbox{{\bf while} $L \neq [ \: ]$ {\bf do}}.
\]

So we can view the {\tt SI} algorithm as a specialization of the {\tt
  GI} algorithm of Section \ref{sec:gen-ite} in which the elements of
the set of functions $G$ (here represented by the list $L$) are
selected in a specific way and in which the $update$ function always
yields the empty set.

In Section \ref{sec:compound} we refined the {\tt GI} algorithm for
the case of compound domains. An analogous refinement of the {\tt SI}
algorithm is straightforward and omitted.  In the next two sections we
show how we can use this refinement of the {\tt SI} algorithm to
derive two well-known constraint propagation algorithms.

\section{{\tt DAC}: a Directional Arc Consistency Algorithm}
\label{sec:directional-arc-algo}

We consider here the notion of directional arc consistency of
Dechter and Pearl \cite{dechter88}.
To derive an algorithm that achieves this local consistency notion we
first characterize it in terms of fixpoints.  To this end, given a
${\cal P}$ and a linear ordering $\prec$ on its variables, we rather
reason in terms of the equivalent CSP ${\cal P}_{\prec}$ obtained from
${\cal P}$ by reordering its variables along $\prec$ so that each
constraint in ${\cal P}_{\prec}$ is on a sequence of variables $x_1,
\LL, x_k$ such that $x_1 \prec x_2 \prec \LL \prec x_k$.

The following characterization holds.

\begin{lemma}[Directional Arc Consistency] \label{lem:darc} 
  Consider a CSP ${\cal P}$ with a linear ordering $\prec$ on its
  variables.  Let ${\cal P}_{\prec} := \p{{\cal C}}{x_1 \in D_1, \LL,
    x_n \in D_n}$.  Then ${\cal P}$ is directionally arc consistent
  w.r.t. $\prec$ iff $(D_1, \LL, D_n)$ is a common fixpoint of the
  functions $\pi^{+}_1$ associated with the binary constraints from
  ${\cal P}_{\prec}$.
\HB
\end{lemma}

We now instantiate in an appropriate way the {\tt SI} algorithm for
compound domains with all the $\pi_1$ functions associated with the
binary constraints from ${\cal P}_{\prec}$.  In this way we obtain an
algorithm that achieves for ${\cal P}$ directional arc consistency
w.r.t. $\prec$.  First, we adjust the definition of semi-commutativity
to functions with different schemes.
To this end consider a sequence of partial orderings $(D_1, \po_1),
\LL , (D_n, \po_n)$ and their Cartesian product $(D, \po )$.  Take
two functions, $f$ with scheme $s$ and $g$ with scheme $t$. We say that $f$
{\em semi-commutes with \/} $g$ ({\em w.r.t. $\po$\/}) if $f^+$ 
semi-commutes with $g^+$ w.r.t. $\po$, that is if
\[
f^+ (g^+ (Q)) \po g^+ (f^+ (Q)).
\]
for all $Q \in D$.

The following lemma is crucial.

\begin{lemma}[Semi-commutativity] \label{lem:semi}
Consider a CSP and two binary constraints of it,
$C_1$ on $u,z$ and $C_2$ on $x,y$, where $y \prec z$.

Then the $\pi_1$ function of $C_1$ semi-commutes with
the $\pi_1$ function of $C_2$
w.r.t. the componentwise  ordering $\supseteq$.
\HB
\end{lemma}

Consider now a CSP ${\cal P}$ with a linear ordering $\prec$ on its
variables and the corresponding CSP ${\cal P}_{\prec}$.
To be able to apply the above lemma we order the $\pi_1$ functions of
the binary constraints of ${\cal P}_{\prec}$ in an appropriate way.
Namely, given two $\pi_1$ functions, $f$ associated with 
a constraint on $u,z$ and $g$ associated with a constraint on $x,y$,
we put $f$ before $g$ if $y \prec z$.

More precisely, let $x_1, \LL, x_n$ be the sequence of the variables
of ${\cal P}_{\prec}$.  So $x_1 \prec x_2 \prec \LL \prec x_n$.
Let for $m \in [1..n]$ the list $L_m$ consist of the $\pi_1$ functions of
those binary constraints of ${\cal P}_{\prec}$ that are
on $x_j, x_m$ for some $x_j$. We order each list $L_m$ arbitrarily.
Consider now the list $L$ resulting from appending $L_n, L_{n-1}, \LL, L_1$,
in that order, so with the elements of $L_n$ in front.
Then by virtue of the Semi-commutativity Lemma
\ref{lem:semi} if the function $f$ precedes the function $g$ in 
the list $L$,
then  $f$ semi-commutes with $g$ w.r.t. the componentwise  
ordering $\supseteq$.

We instantiate now the refinement of the {\tt SI} algorithm for the
compound domains by the above-defined list $L$ and each $\bot_i$ equal
to the domain $D_i$ of the variable $x_i$. We assume that $L$ has $k$
elements.  We obtain then the following algorithm.
\II

\NI
{\sc Directional Arc Consistency Algorithm ({\tt DARC})}
\begin{tabbing}
\= $d := (\bot_1, \LL, \bot_n)$; \\
\> {\bf for} $i := 1$ {\bf to} $k$ {\bf do} \\
\> \qquad $g := {\bf head}(L)$; suppose $g$ is with scheme $s$; \\
\> \qquad $L := {\bf tail}(L)$; \\
\> \qquad $d[s] := g(d[s])$ \\
\> {\bf od} 
\end{tabbing}
\NI

This algorithm enjoys the following properties.

\begin{theorem}[{\tt DARC} Algorithm] \label{thm:darc-algo}
Consider a CSP ${\cal P}$ with a linear ordering $\prec$ on its
variables.  Let 
${\cal P}_{\prec} := \p{{\cal C}}{x_1 \in D_1, \LL, x_n \in D_n}$. 

The {\tt DARC} algorithm always terminates.  
Let ${\cal P'}$ be the CSP determined by ${\cal P}_{\prec}$ and
the sequence of the domains $D'_1, \LL,  D'_n$ 
computed in $d$. Then
\begin{enumerate}\smallromani

\item ${\cal P'}$ is the $\sqsubseteq_d$-least CSP in
$\C{{\cal P}_1 \mid {\cal P}_{\prec} \sqsubseteq_d {\cal P}_1}$
that is directionally arc consistent w.r.t. $\prec$,

\item ${\cal P'}$ is equivalent to ${\cal P}$.
\HB
\end{enumerate}
\end{theorem}

Note that in contrast to the {\tt HYPER-ARC} Algorithm Theorem
\ref{thm:hyper-arc} we do not need to assume here that each domain is
finite.

Assume now that for each pair of variables $x,y$ of the original CSP
${\cal P}$ there exists precisely one constraint on $x,y$. The same
holds then for ${\cal P}_{\prec}$.  Suppose that 
${\cal P}_{\prec} := \p{{\cal C}}{x_1 \in D_1, \LL, x_n \in D_n}$. 
Denote the
unique constraint of ${\cal P}_{\prec}$ on $x_i, x_j$ by $C_{i,j}$.
The above {\tt DARC} algorithm can then be rewritten as the following
algorithm known as the {\tt DAC} algorithm
of Dechter and Pearl \cite{dechter88}:
\II

\NI
\begin{tabbing}
\= {\bf for} $j := n$ {\bf to} $2$ {\bf by} $-1$ {\bf do} \\
\> \qquad {\bf for} $i := 1$ {\bf to} $j-1$ {\bf do} \\
\> \qquad \qquad $D_i := \C{a \in D_i \mid \te \: b \in D_j \: (a,b) \in C_{i,j}}$ \\
\> \qquad {\bf od} \\
\> {\bf od} 
\end{tabbing}

\section{{\tt DPC}: a Directional Path Consistency Algorithm}
\label{sec:directional-path-algo}

In this section we deal with the notion of directional path
consistency defined in Dechter and Pearl \cite{dechter88}.  As before
we first characterize this local consistency notion in terms of
fixpoints.  To this end, as in the previous section, given a
normalized CSP ${\cal P}$ we rather consider the equivalent CSP ${\cal
  P}_{\prec}$.  The variables of ${\cal P}_{\prec}$ are ordered
according to $\prec$ and on each pair of its variables there exists a
unique constraint.

The following is a counterpart of the
Directional Arc Consistency Lemma \ref{lem:darc}.

\begin{lemma}[Directional Path Consistency] \label{lem:dpath}
  Consider a normalized CSP ${\cal P}$ with a linear ordering $\prec$ on its
  variables.  Let ${{\cal P}_{\prec}} := \p{{C_1, \LL,
      C_k}}{{\cal DE}}$.  Then ${\cal P}$ is directionally path
  consistent w.r.t. $\prec$ iff
$(C_1, \LL, C_k)$ is a common fixpoint of all
functions $(f^{z}_{x,y})^{+}$ associated with
the subsequences $x,y,z$ of the variables of ${{\cal P}_{\prec}}$.
\HB
\end{lemma}

To obtain an algorithm that achieves directional path consistency we
now instantiate in an appropriate way the {\tt SI} algorithm.  To this
end we need the following lemma.  

\begin{lemma}[Semi-commutativity] \label{lem:semip}
Consider a normalized CSP
and two subsequences of its variables,
$x_1 ,y_1 ,z$ and  $x_2, y_2 , u$.
Suppose that $u \prec z$.

Then the function $f^{z}_{x_1 ,y_1}$ semi-commutes with
the function $f^{u}_{x_2, y_2}$
w.r.t. the componentwise  ordering $\supseteq$.
\HB
\end{lemma}

Consider now a normalized CSP ${\cal P}$ with a linear ordering
$\prec$ on its variables and the corresponding CSP ${{\cal
    P}_{\prec}}$.  To be able to apply the above lemma we order in an
appropriate way the $f^{t}_{r,s}$ functions, where the variables
$r,s,t$ are such that $r \prec s \prec t$. Namely, we put $f^{z}_{x_1
  ,y_1}$ before $f^{u}_{x_2, y_2}$ if $u \prec z$.

More precisely, let $x_1, \LL, x_n$ be the sequence of the variables
of ${\cal P}_{\prec}$, that is $x_1 \prec x_2 \prec \LL \prec x_n$.
Let for $m \in [1..n]$ the list $L_m$ consist of the functions
$f^{x_m}_{x_i, x_j}$ for some $x_i$ and $x_j$.  We order each list
$L_m$ arbitrarily and consider the list $L$ resulting from appending
$L_n, L_{n-1}, \LL, L_1$, in that order.  Then by virtue of the
Semi-commutativity Lemma \ref{lem:dpath} if the function $f$ precedes
the function $g$ in the list $L$, then $f$ semi-commutes with $g$
w.r.t. the componentwise ordering $\supseteq$.

We instantiate now the refinement of the {\tt SI} algorithm for the
compound domains by the above-defined list $L$ and each $\bot_i$ equal
to the constraint $C_i$. We assume that $L$ has $k$ elements.  This
yields the {\sc Directional Path Consistency Algorithm ({\tt DPATH})}
that, apart from of the different choice of the constituent partial
orderings, is identical to the {\sc Directional Arc Consistency
  Algorithm} {\tt DARC} of the previous section.
Consequently, the {\tt DPATH} algorithm enjoys analogous properties
as the {\tt DARC} algorithm. They are summarized in the following theorem.

\begin{theorem}[{\tt DPATH} Algorithm] \label{thm:Parc-algo}
Consider a CSP ${\cal P}$ with a linear ordering $\prec$ on its
variables.  Let
${{\cal P}_{\prec}} := \p{{C_1, \LL, C_k}}{{\cal DE}}$. 

The {\tt DPATH} algorithm always terminates.
Let ${\cal P'} := \p{{C'_1, \LL, C'_k}}{{\cal DE}}$,
where the sequence of the constraints $C'_1, \LL,  C'_k$ is
computed in $d$. Then
\begin{enumerate}\smallromani

\item ${\cal P'}$ is the $\sqsubseteq_c$-least CSP in
$\C{{\cal P}_1 \mid {{\cal P}_{\prec}} \sqsubseteq_d {\cal P}_1}$
that is directionally path consistent w.r.t. $\prec$,

\item ${\cal P'}$ is equivalent to ${\cal P}$.
\HB
\end{enumerate}
\end{theorem}

As in the case of the {\tt DARC} Algorithm Theorem \ref{thm:darc-algo}
we do not need to assume here that each domain is finite.

Assume now that that $x_1, \LL, x_n$ is the sequence of the variables
of ${{\cal P}_{\prec}}$.  Denote the unique constraint of
${\cal P}_{\prec}$ on $x_i, x_j$ by $C_{i,j}$.

The above {\tt DPATH} algorithm can then be rewritten  as the following
algorithm known as the {\tt DPC} algorithm
of Dechter and Pearl \cite{dechter88}:
\II

\NI
\begin{tabbing}
\= {\bf for} $m := n$ {\bf to} $3$ {\bf by} $-1$ {\bf do} \\
\> \qquad {\bf for} $j := 1$ {\bf to} $m-1$ {\bf do} \\
\>  \qquad \qquad {\bf for} $i := 1$ {\bf to} $j-1$ {\bf do} \\
\>  \qquad \qquad \qquad $C_{i,j} := C_{i,m} \cdot C^T_{j,m}$ \\
\>  \qquad \qquad {\bf od} \\
\> \qquad {\bf od} \\
\> {\bf od}
\end{tabbing}

\section{Conclusions}

\label{sec:conclusions}

In this article we introduced a general framework for constraint
propagation. It allowed us to present and explain various constraint
propagation algorithms in a uniform way.  Using such a single
framework we can easier verify, compare, modify, parallelize or
combine these algorithms. The last point has already been made to
large extent in Benhamou \cite{Ben96}.  Additionally, we clarified the
role played by the notions of commutativity and semi-commutativity.

The line of research presented here could be extended in a
number of ways. First, it would be interesting to find examples of
existing constraint propagation algorithms that could be improved by
using the notions of commutativity and semi-commutativity.

Second, as already stated in Apt \cite{Apt99b}, it would be useful to
explain in a similar way other constraint propagation algorithms such
as the {\tt AC-4} algorithm of Mohr and Henderson \cite{MH86}, the
{\tt PC-4} algorithm of Han and Lee \cite{HL88}, or the {\tt GAC-4}
algorithm of Mohr and Masini \cite{MM88}.  The complication is that
these algorithms operate on some extension of the original CSP.

Finally, it would be useful to apply the approach of this paper to
derive constraint propagation algorithms for the semiring-based
constraint satisfaction framework of Bistarelli, Montanari and Rossi
\cite{BMR97} that provides a unified model for several classes of
``nonstandard'' constraints satisfaction problems.


\begin{thebibliography}{10}

\bibitem{Apt99b}
K.~R. Apt.
\newblock The essence of constraint propagation.
\newblock {\em Theoretical Computer Science}, 221(1--2):179--210, 1999.
\newblock Available via \verb+http://xxx.lanl.gov/archive/cs/+.

\bibitem{Ben96}
F.~Benhamou.
\newblock Heterogeneous constraint solving.
\newblock In M.~Hanus and M.~Rodriguez-Artalejo, editors, {\em Proceeding of
  the Fifth International Conference on Algebraic and Logic Programming (ALP
  96)}, Lecture Notes in Computer Science 1139, pages 62--76, Berlin, 1996.
  Springer-Verlag.

\bibitem{BO97}
F.~Benhamou and W.~Older.
\newblock Applying interval arithmetic to real, integer and {B}oolean
  constraints.
\newblock {\em Journal of Logic Programming}, 32(1):1--24, 1997.

\bibitem{BMR97}
S.~Bistarelli, U.~Montanari, and F.~Rossi.
\newblock Semiring-based constraint satisfaction and optimization.
\newblock {\em Journal of the ACM}, 44(2):201--236, March 1997.

\bibitem{D99}
R.~Dechter.
\newblock Bucket elimination: A unifying framework for structure-driven
  inference.
\newblock {\em Artificial Intelligence}, 1999.
\newblock To appear.

\bibitem{dechter88}
R.~Dechter and J.~Pearl.
\newblock Network-based heuristics for constraint-satisfaction problems.
\newblock {\em Artificial Intelligence}, 34(1):1--38, January 1988.

\bibitem{DvB97}
R.~Dechter and P.~van Beek.
\newblock Local and global relational consistency.
\newblock {\em Theoretical Computer Science}, 173(1):283--308, 20~February
  1997.

\bibitem{HL88}
C.~Han and C.~Lee.
\newblock Comments on {M}ohr and {H}enderson's path consistency algorithm.
\newblock {\em Artificial Intelligence}, 36:125--130, 1988.

\bibitem{mackworth-consistency}
A.~Mackworth.
\newblock Consistency in networks of relations.
\newblock {\em Artificial Intelligence}, 8(1):99--118, 1977.

\bibitem{MS98b}
K.~Marriott and P.~Stuckey.
\newblock {\em Programming with Constraints}.
\newblock The MIT Press, Cambridge, Massachusetts, 1998.

\bibitem{MH86}
R.~Mohr and T.C. Henderson.
\newblock Arc-consistency and path-consistency revisited.
\newblock {\em Artificial Intelligence}, 28:225--233, 1986.

\bibitem{MM88}
R.~Mohr and G.~Masini.
\newblock Good old discrete relaxation.
\newblock In Y.~Kodratoff, editor, {\em Proceedings of the 8th European
  Conference on Artificial Intelligence (ECAI)}, pages 651--656. Pitman
  Publishers, 1988.

\bibitem{MR99}
E.~Monfroy and J.-H. {R{\'{e}}ty}.
\newblock Chaotic iteration for distributed constraint propagation.
\newblock In J.~Carroll, H.~Haddad, D.~Oppenheim, B.~Bryant, and G.~Lamont,
  editors, {\em Proceedings of The 1999 ACM Symposium on Applied Computing,
  SAC'99}, pages 19--24, San Antonio, Texas, USA, March 1999. ACM Press.

\bibitem{montanari-networks}
U.~Montanari.
\newblock Networks of constraints: Fundamental properties and applications to
  picture processing.
\newblock {\em Information Science}, 7(2):95--132, 1974.
\newblock Also Technical Report, Carnegie Mellon University, 1971.

\bibitem{saraswat-semantic}
V.A. Saraswat, M.~Rinard, and P.~Panangaden.
\newblock Semantic foundations of concurrent constraint programming.
\newblock In {\em Proceedings of the Eighteenth Annual ACM Symposium on
  Principles of Programming Languages {(POPL'91)}}, pages 333--352, 1991.

\bibitem{TU96}
V.~Telerman and D.~Ushakov.
\newblock Data types in subdefinite models.
\newblock In J.~A.~Campbell J.~Calmet and J.~Pfalzgraf, editors, {\em
  Artificial Intelligence and Symbolic Mathematical Computations}, Lecture
  Notes in Computer Science 1138, pages 305--319, Berlin, 1996.
  Springer-Verlag.

\bibitem{Tsa93}
E.~Tsang.
\newblock {\em Foundations of Constraint Satisfaction}.
\newblock Academic Press, 1993.

\end{thebibliography}

\end{document}